\documentclass[12pt]{article}
\usepackage{graphicx} 
\usepackage{basicreq}
\usepackage{multirow}
\usepackage{amsmath,amssymb,amsfonts}
\usepackage{amsthm}
\usepackage{mathrsfs}
\usepackage[title]{appendix}
\usepackage{xcolor}
\usepackage{textcomp}
\usepackage{manyfoot}
\usepackage{booktabs}
\usepackage{algorithm}
\usepackage{algorithmicx}
\usepackage{algpseudocode}
\usepackage{listings}
\usepackage{multirow}
\usepackage{hhline}
\usepackage{rotating}
\usepackage{hyperref}
\usepackage{layout}
\usepackage{geometry}
\usepackage{subcaption}

\title{Navigation of micro-robot swarms for targeted delivery using reinforcement learning}
\author{Akshatha Jagadish, Manoj Varma }
\date{}

\begin{document}

\maketitle

\setlength{\oddsidemargin}{10pt}
\abstract{Micro-robotics is quickly emerging to be a promising technological solution  to many medical treatments with focus on targeted drug delivery. They are effective when working in swarms whose individual control is mostly infeasible owing to their minute size. Controlling a number of robots with a single controller is thus important and artificial intelligence can help us perform this task successfully. In this work, we use the Reinforcement Learning (RL) algorithms Proximal Policy Optimization (PPO) and Robust Policy Optimization (RPO) to navigate a swarm of 4,9 and 16 micro-swimmers under hydro-dynamic effects, controlled by their orientation, towards a circular absorbing target. We look at both PPO and RPO's performances with limited state information scenarios and also test their robustness for random target location and size. We use curriculum learning to improve upon the performance and demonstrate the same on learning to navigate a swarm of 25 swimmers and steering the swarm to exemplify the manoeuvring capabilities of the RL model.}

\textbf{Keywords}: micro-swimmers, RL, PPO, RPO, curriculum learning, swarm-control

\section{Introduction}
\label{sec:IntroCh6}

Micro-scale is a fertile area for research and provides the promise of great applications in a variety of fields such as micro-surgery \cite{Li17}, micro-manufacturing \cite{Goodrich17}, cargo delivery \cite{Yang20}, pollution rectification \cite{Soler13,Chen21} and many more. While there has been substantial research going on to understand the physics at this scale for many decades, the research in the design and development of robots that can operate at this scale has exponentially increased in recent years, and we see different methods of realizing them \cite{Li16RL,Li17,Yin21}. In addition to the design and propulsion methods, researchers have also been looking at different navigation strategies for these micro-robots \cite{Ider21}. These methods, however, require complete information of the environment that the micro-robots operate in, which is generally difficult to obtain. 

The physical system of micro-robots can be controlled computationally, making it a cyber-physical system, which makes it scalable and reliable. Here, we explore reinforcement learning (RL) as the computational part of the system owing to its incredible performance in recent years in different fields of engineering, such as games \cite{Vinyals19, Mnih15}, robotics \cite{Kober13, breyer19, Kul19, kalashnikov18qtopt}, operations \cite{Waschneck18}, finance \cite{liu21finrl} and healthcare \cite{Yu21}. 

Reinforcement learning is a type of machine learning (ML) technique where an agent learns through interaction with the environment. The RL agent starts choosing actions by trial and error and gradually learns from the rewards of its actions in the environment. Some environments are hard to model because of its complex dynamics and multiple parameters affecting its state. RL can perform well even in these model-free environments and is thus suitable for our application. There are numerous algorithms developed to implement RL, from which we choose Proximal Policy Optimization (PPO) \cite{Schulman17} and Robust Policy Optimization (RPO) \cite{Rahman23} to be the most appropriate for the task of guiding micro-robotic swarm because of their performance in model-free, continuous action space environments. RL has been proved useful for such navigation strategies of micro-robots recently \cite{Yang19,Falk21,Muinos21,Yang20RL}. We later show that curriculum learning helps in improving the performance of the already better-performing RPO.

In our work, we try to navigate a swarm of magnetically controlled helical micro-swimmers to reach a target aided by RL algorithm. We use a simulated model of micro-swimmers which acts as the environment for the RL agent to operate in. The RL agent acts as the global controller of the entire swarm of micro-swimmers. It gets the current state of the entire swarm and environment and outputs the action that needs to be taken by the magnetic controller in order to navigate the swarm of micro-swimmers. In our experiment, we use PPO and RPO as the RL algorithms due to their suitability for the required application, as explained in section \ref{subsubsec:algosel_ch6}.

The chapter is organised as follows. In section \ref{subsec:model_ch6}, we explain the micro-swimmer model. In section \ref{subsec:RL_details_ch6}, we describe the RL algorithm details like state, action, rewards in section \ref{subsubsec:state_action_ch6} and choosing the algorithm in section \ref{subsubsec:algosel_ch6}. In section \ref{sec:exp_and_result_Ch6}, we look at the experimental results of different scenarios of the environment with increasing complexity. Finally, we conclude in section \ref{sec:conclusion_ch6}.

\section{Methods}
\label{sec:methods_ch6}
\subsection{Simulation model}
\label{subsec:model_ch6}
Here, we describe the two dimensional simulation model of the environment of the swarm of micro-swimmers. The micro-swimmers are modelled after rigid helical swimmers. These artificial micro-swimmers are controlled using rotating magnetic field generated by triaxial Helmholtz coils \cite{Ghosh09}. This magnetic field aligns the helical swimmers and rotates them around their helical axis causing linear motion along their axial direction. We model these dynamics using the equation \ref{eqn:ch6_1} for each swimmer $i$.

\begin{equation}
\label{eqn:ch6_1}
\begin{aligned}
\Delta x_i = v \Delta t cos(\theta_i)\\
\Delta y_i = v \Delta t sin(\theta_i)  
\end{aligned}
\end{equation}

Where $\Delta x$ and $\Delta y$ are the positional increments of the microswimmer along $x$ and $y$ directions respectively, $\Delta t$ is the unit simulation time and $v$ is the linear velocity of the swimmer. Note that $v$ is set by the Helmholtz coil. $v$ is proportional to the frequency of the rotating magnetic field. As the frequency increases, the number of rotations of the swimmer increases, as a result of which the swimmer moves faster, i.e., the velocity increases. 
$\theta_i$ is the angle of orientation along which the velocity is applied. It incorporates two components, as shown in equation \ref{eqn:ch6_2}. 

\begin{equation}
\label{eqn:ch6_2}
\theta_i = \rho_i \theta_{hyd} + (1-\rho_i)\theta_m
\end{equation}

Where $\theta_m$ is the actual orientation of the swimmers set by the Helmholtz coil. $\theta_{hyd}$ is the orientation due to the hydrodynamic effect of the surroundings, and it is taken as $\theta_{hyd} = \theta_m \text{\textendash} \:90^{\circ}$ because of the transverse drift due to the fluid flow of the nearby micro-swimmers \cite{Pal22}. $\rho_i$ denotes the weight of the hydrodynamic effect and is described by equation \ref{eqn:ch6_3}. It is also ceiled at the maximum value of 1.

\begin{equation}
\label{eqn:ch6_3}
\rho_i = \sum_{\forall j, j \neq i} (2/r_{ij}^2)  
\end{equation}

The target region where the micro-swimmer swarm is to be navigated is assumed to be circular and specified by $(x_t, y_t, r_t)$ where $x_t, y_t$ denotes the centre in the 2d space and $r_t$ is the radius of the target region. The target is absorbing in nature, which means that the micro-swimmers reaching the target get stuck there and do not move again.

Also, note that the $\xi_x$ and $\xi_y$ that are usually considered in the Langevin model of micro-robots which capture the Brownian motion \cite{Bechinger16}, are not considered here because the effect of the magnetic field of the Helmholtz coil is high enough to suppress the thermal effects at low Reynold’s number \cite{Purcell77}.

In our simulations, we keep the frequency and hence the velocity of the micro-swimmers constant, and we only learn the orientation $\theta_m$ through RL to navigate the swarm of robots. The challenge here lies in controlling multiple swimmers with a single control parameter in the presence of the hydrodynamic effect.

\subsection{Formulating RL}
\label{subsec:RL_details_ch6}
Here, we describe the details of the RL algorithm in connection with the simulation model described in the previous sub-section.

\subsubsection{Defining state, action and reward}
\label{subsubsec:state_action_ch6}
Reinforcement learning consists of two main components: the agent and the environment. These two components interact with each other using three quantities, namely states, actions and rewards. 

The state is the description of the environment given to the RL agent. For our problem of navigating the microswimmers to a target region, the essential state information is the state information of the swarm of micro-swimmers and the state of the target as shown in equation \ref{eqn:ch6:4}. 

\begin{equation}
\label{eqn:ch6:4}
S = (S_{swimmers}, S_{target})
\end{equation}

$S_{target}$ describes the target state, and as mentioned in the environment description, $S_{target} = (x_t, y_t, r_t)$, which completely defines the target. $S_{swimmers}$ can be done in multiple ways. For instance, it can either be the 2d positional coordinates of all the swimmers along with their orientation information or just the positional information or just the mean positional information. We run the RL simulations for all these states to understand their performance in complete and limited state information, as will be later described in the results section.

Action is the control parameter value the agent calculates and sends to the environment. In our problem, $\theta_m$, as specified in section \ref{subsec:model_ch6} denotes the action which orients the swimmers in the required direction.

The reward is the feedback the RL agent receives from the environment upon execution of the action that the RL agent determined. The agent modifies its action determination policy by analyzing this reward. We use the number of swimmers reaching the target as the reward for our navigation problem, which is the goal for our RL simulations and the quantity that needs to be maximized. 

\subsubsection{Algorithm selection}
\label{subsubsec:algosel_ch6}
There are many algorithms available for implementing RL. We wanted to choose one suitable for a model-free environment that works with continuous state and action spaces. We found that algorithms like actor-critic and trust region policy optimization are ideal for our application as they focus on policy optimization directly. Proximal policy optimization (PPO), being state of the art in this class of algorithms, uses the suitable characteristics from both the algorithms mentioned above. We thus choose PPO for our current problem statement.

PPO takes the advantages of actor-critic algorithms along with trust region optimization and minibatch updates. Actor-critic algorithms consist of two parts: actor and critic, where the former chooses the action to be taken, and the latter evaluates the performance of that action. The two optimize each other using separate loss functions and are neural networks for PPO implementation. The second important characteristic of PPO is trust region optimization, where the policy being learnt is not allowed to change significantly in a single update step. This is done by limiting the KL divergence between the old and new policies. The third characteristic, minibatch updates, is an implementation strategy of applying sample efficiency to the algorithm. This enables the RL setup to learn from multiple samples simultaneously rather than just a single sample, like in the predecessor RL algorithm, TRPO (Trust Region Policy Optimization). This last characteristic helps learn the policy quicker than previous policy gradient methods in RL \cite{Schulman15}.

We came across another algorithm called Robust Policy Optimization (RPO), which achieves increased exploration capabilities in the PPO algorithm with just a minor modification in the implementation \cite{Rahman23}. We use the cleanRL implementations \cite{Huang22} with few modifications for RL simulations to suit our micro-swimmer navigation environment. CleanRL provides single page code for each RL algorithm and incorporates features like state normalization and clipping, action clipping, and generalized advantage estimation that help in better convergence of policy estimation through RL.

\begin{figure}
    \centering
    \includegraphics[width=0.7\textwidth]{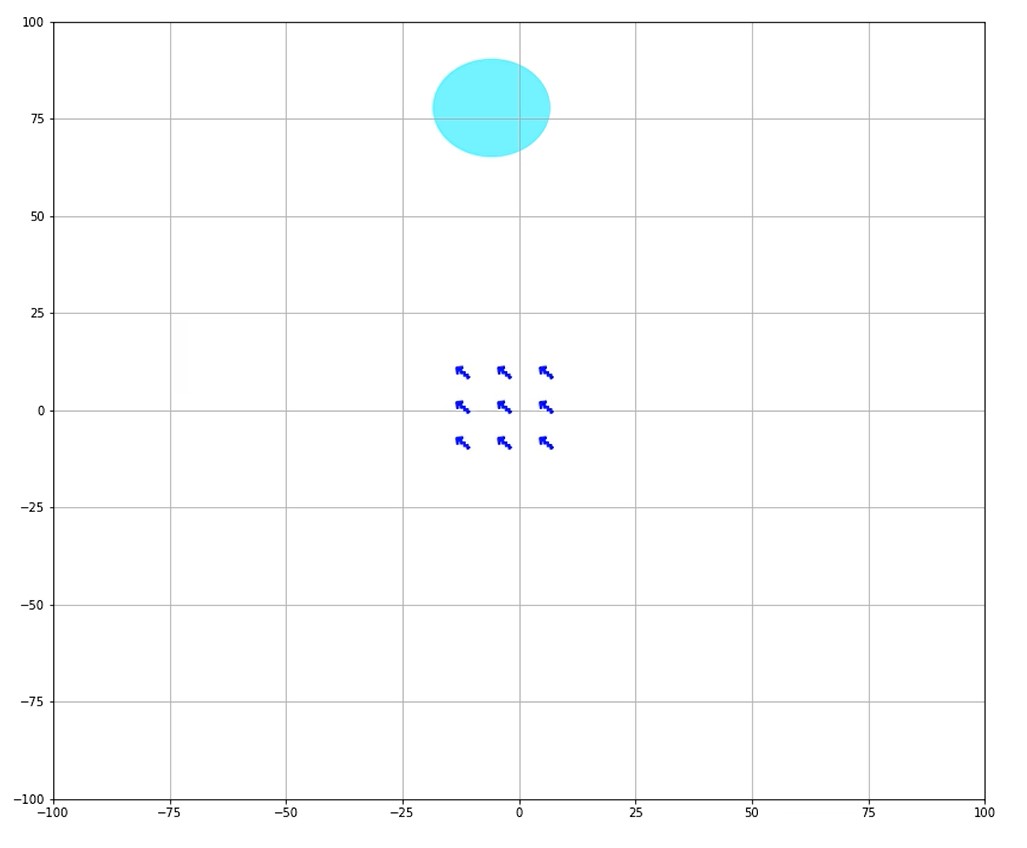}
    \caption{Sample environment setup at the beginning of anepisode}
    \label{fig:env_setup_ch6}
\end{figure}

\begin{figure}[t]
    \centering
    \begin{subfigure}{0.35\textwidth}
    \centering
    \frame{\includegraphics[width=\textwidth]{ 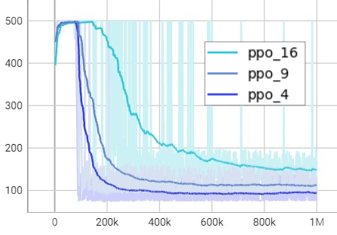}}
    \caption{}
    \label{fig:0_ppo_len_ch6}
    \end{subfigure}\hspace{0.5cm}
    \begin{subfigure}{0.35\textwidth}
    \centering
    \frame{\includegraphics[width=\textwidth]{ 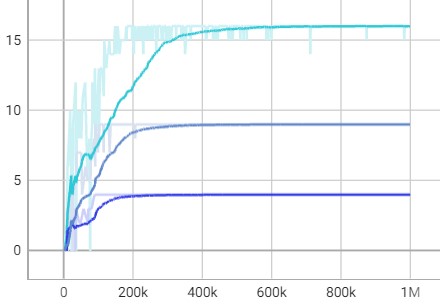}}
    \caption{}
    \label{fig:0_ppo_ret_ch6}
    \end{subfigure}\hfill
    \begin{subfigure}{0.35\textwidth}
    \centering
    \frame{\includegraphics[width=\textwidth]{ 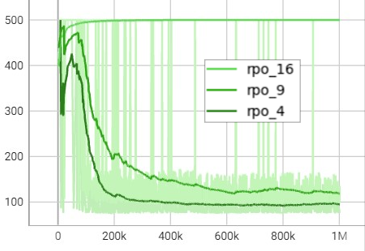}}
    \caption{}
    \label{fig:0_rpo_len_ch6}
    \end{subfigure}\hspace{0.5cm}
    \begin{subfigure}{0.35\textwidth}
    \centering
    \frame{\includegraphics[width=\textwidth]{ 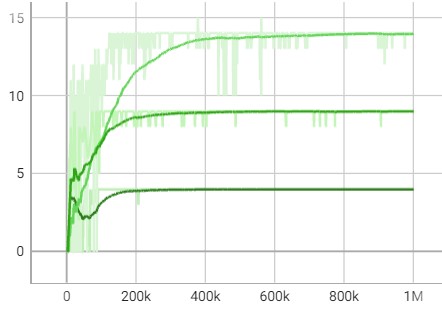}}
    \caption{}
    \label{fig:0_rpo_ret_ch6}
    \end{subfigure}\hfill
    \caption{Performance of primary experiment for PPO and RPO with respect to time-steps for 4, 9 and 16 swimmers: (a) episodic lengths of PPO (b) episodic returns of PPO (c) episodic lengths of RPO (d) episodic returns of RPO }
    \label{fig:primary_expt_ch6}
\end{figure}

\section{RL Simulations and Results}
\label{sec:exp_and_result_Ch6}
\subsection{Environmental set-up}
Here, we describe the implementation details of the simulation environment of the swarm of micro-swimmers. The micro-swimmer simulation environment is set-up with the following parameters. The swimmers are initially located in a square shape with the initial mean position located at the centre of the 2d-coordinate system as shown in figure \ref{fig:env_setup_ch6}. The swimmers are represented by dark blue squiggly arrows depicting the spiral microswimmers. The target is represented by the light blue circle. The spacing between the swimmers is around 6um. Their orientation is chosen randomly for every episode starting. The target centre is randomly selected within a 100um radius from the initial mean position of the micro-swimmers. The target radius is randomly selected within 5-20um. We vary the number of swimmers to 4, 9 and 16 for all the simulations. 

An episode is defined as the beginning of a simulation run to the end of the run. The termination of the episode is when all the swimmers reach the target or the length of the episode runs beyond 500 time-steps or the mean position of micro-swimmers is more than 200um from the target centre, where the swimmers have wandered too far from the target. The velocity induced from the Helmholtz coil to the micro-swimmers is taken to be 10um/sec and the orientation is given by the RL agent at each time step. Every time step is a duration of 0.1s.

\renewcommand{\arraystretch}{1.5}
\begin{table}[ht!]
\caption{Performance: smoothened return values after learning}
\label{table:performance_ch6}
\centering
\begin{tabular}{|c|c|c|c|c|}
\hline
Environment & Description &\begin{tabular}[c]{@{}c@{}}no. of \\ swimmers\end{tabular} & PPO & RPO \\ \hline
\multirow{3}{*}{env-0} & \multirow{3}{*}{\begin{tabular}[c]{@{}c@{}}Constant target location and size \\ with full state information\end{tabular}} & 4 & 4 & 4 \\ \hhline{~~---} 
 & & 9 & 9 & 8.9 \\ \hhline{~~---} 
 & & 16 & 16 & 13.9 \\ \hline
\multirow{3}{*}{env-1a} & \multirow{3}{*}{\begin{tabular}[c]{@{}c@{}}Swimmer positional state information \\ without orientation information\end{tabular}} & 4 & 3.9 & 3.7 \\ \hhline{~~---} 
 & & 9 & 8.6 & 8.5 \\ \hhline{~~---} 
 & & 16 & 15.6 & 15.1 \\ \hline
\multirow{3}{*}{env-1b} & \multirow{3}{*}{\begin{tabular}[c]{@{}c@{}}Swimmer mean position and \\ orientation information\end{tabular}} & 4 & 4 & 3.9 \\ \hhline{~~---} 
 & & 9 & 7.8 & 8.9 \\ \hhline{~~---} 
 & & 16 & 11.8 & 15.9 \\ \hline
\multirow{3}{*}{env-1c} & \multirow{3}{*}{\begin{tabular}[c]{@{}c@{}}Mean position and orientation \\information of swimmers \\outside target\end{tabular}} & 4 & 4 & 4 \\ \hhline{~~---} 
 & & 9 & 9 & 8.9 \\ \hhline{~~---} 
 & & 16 & 11.9 & 11.9 \\ \hline
\multirow{3}{*}{env-2} & \multirow{3}{*}{\begin{tabular}[c]{@{}c@{}}Random target location and size \\ with full state information\end{tabular}} & 4 & 0.2 & 3.9 \\ \hhline{~~---} 
 & & 9 & 0.9 & 8.5 \\ \hhline{~~---} 
 & & 16 & 2.7 & 14.1 \\ \hline
env-2-om & \begin{tabular}[c]{@{}c@{}}Multiple environments with \\target orientation included in \\ state information on env-2\end{tabular} & 16 & - & 14.6 \\ \hline
\multirow{2}{*}{env-2-omc} & \multirow{2}{*}{\begin{tabular}[c]{@{}c@{}}Curriculum learning on env-2om\end{tabular}} & 16 & - & 15.8 \\ \hhline{~~---} 
 & & 25 & - & 24.5 \\ \hline
\end{tabular}
\end{table}

The reward at each time step is calculated as the number of swimmers reaching the target at that time step. The return is the weighted cumulative reward given by $E\sum_{n=1}^{\infty}\gamma^n[R(s_n)]$ where gamma is the discount factor taken to be 0.99 to specify to the agent to focus on distant rewards or ultimate task at hand. 
The neural networks of the actor and critic contain a hidden layer of 64 neurons along with an initial layer and a final layer. The final layer for the actor provides the mean and standard deviation which provides the normal distribution description of the action that needs to be taken. The final layer of the critic provides the estimate of the value function which is the expected value of the episodic return. The other parameters like the number of mini-batches, learning rate and total time steps are all kept the default values taken from the cleanRL code. 

The following subsections describe the different experiments performed and their corresponding performances, a summary of which is presented in the table \ref{table:performance_ch6}.

\subsection{Primary experiment}
Here, we performed the learning of navigation of micro-swimmers towards a constant target (constant location and size). This experiment was performed for 4, 9 and 16 micro-swimmers, where the RL agent was given complete state information of the positions and orientations of all swimmers in addition to target location and size, in each case. Thus the size of the state vector was 15, 30 and 51 respectively. 

Figure \ref{fig:primary_expt_ch6} represents the episodic return and length across the time steps for 4, 9 and 16 swimmers for both PPO and RPO algorithms. We observe that both the algorithms learn well and  that the experiments with the higher number of swimmers take more time to converge compared to the lower number of swimmers. RPO for 16 swimmers is stuck at a local minimum but is not yet stable at 1 million time steps. We see that it converges to the full expected return if run for more time steps. PPO learns very well for this experiment as seen in the table \ref{table:performance_ch6} and figure \ref{fig:0_ppo_ret_ch6} for all 4, 9 and 16 swimmers.

\begin{figure}[t]
    \centering
    \begin{subfigure}{0.3\textwidth}
    \centering
    \frame{\includegraphics[width=\textwidth]{ 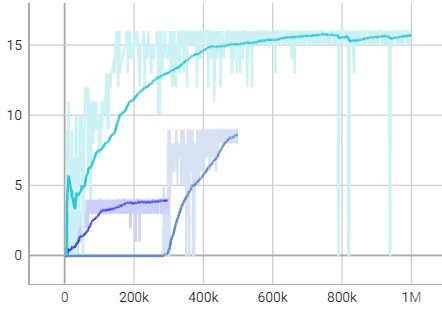}}
    \caption{}
    \label{fig:1a_ppo_ch6}
    \end{subfigure}\hfill
    \begin{subfigure}{0.3\textwidth}
    \centering
    \frame{\includegraphics[width=\textwidth]{ 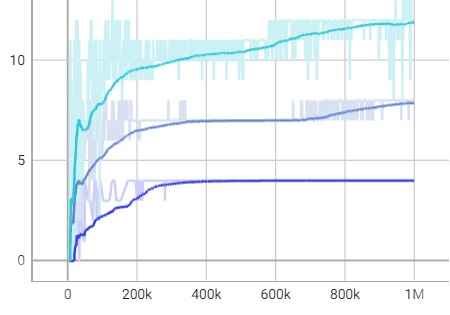}}
    \caption{}
    \label{fig:1b_ppo_ch6}
    \end{subfigure}\hfill
    \begin{subfigure}{0.3\textwidth}
    \centering
    \frame{\includegraphics[width=\textwidth]{ 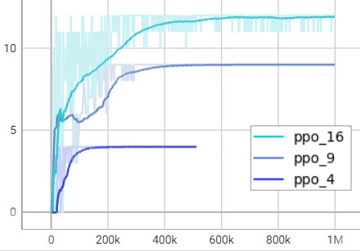}}
    \caption{}
    \label{fig:1c_ppo_ch6}
    \end{subfigure}\hfill
    \begin{subfigure}{0.3\textwidth}
    \centering
    \frame{\includegraphics[width=\textwidth]{ 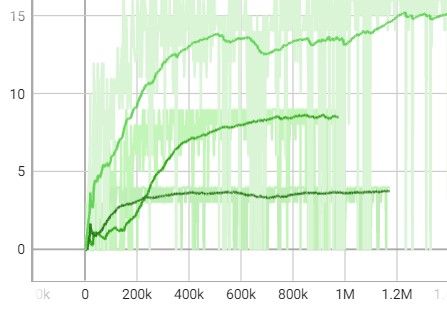}}
    \caption{}
    \label{fig:1a_rpo_ch6}
    \end{subfigure}\hfill
    \begin{subfigure}{0.3\textwidth}
    \centering
    \frame{\includegraphics[width=\textwidth]{ 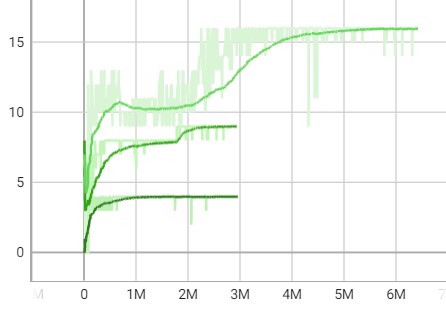}}
    \caption{}
    \label{fig:1b_rpo_ch6}
    \end{subfigure}\hfill
    \begin{subfigure}{0.3\textwidth}
    \centering
    \frame{\includegraphics[width=\textwidth]{ 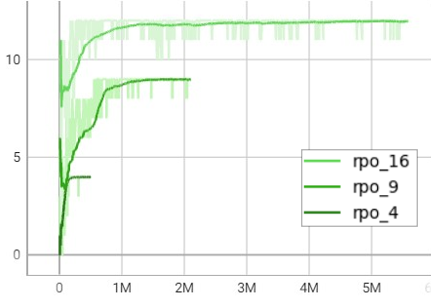}}
    \caption{}
    \label{fig:1c_rpo_ch6}
    \end{subfigure}\hfill
    \caption{Performance of state modification experiments for PPO and RPO in terms of episodic returns with respect to time-steps for 4, 9 and 16 swimmers: (a) PPO with env-1a (b) PPO with env-1b (c) PPO with env-1c (d) RPO with env-1a (e) RPO with env-1b (f) RPO with env-1c}
    \label{fig:state_modification_ch6}
\end{figure}

\subsection{State information modification experiments}
Here we performed 3 sets of experiments again for 4, 9 and 16 swimmers, the results of which are shown in figure \ref{fig:state_modification_ch6}. The same seed is considered for all experiments with target specification: $(10.56,41.63,9.36)\mu m$, the same as for primary experiment.

In the first set, we provided only the position information of swimmers in $s_{swimmers}$ and no orientation information. Thus, the size of state vector was 11, 21 and 35 respectively for 4, 9 and 16 swimmers. Figures \ref{fig:1a_ppo_ch6} and \ref{fig:1a_rpo_ch6} show the performance of the learning activity with episodic return over the time steps for PPO and RPO respectively. We observe they are relatively slower to converge compared to the primary case because of less information but they do converge.

In the second set, we provided the mean position and orientation information of swimmers in $s_{swimmers}$. Thus, the size of state vector was 6 for any number of swimmers. In this case, the figures \ref{fig:1b_ppo_ch6} and \ref{fig:1b_rpo_ch6} show the performance evolution of the learning environment for PPO and RPO respectively. Here, we observe that both PPO and RPO converge for 4 swimmers but are stuck at local minimum for 9 and 16 swimmers. Probably, the mean information of all the swimmers is useless once most of the swimmers reach the target. RPO shows relatively faster learning compared to PPO.

In the final set of the state information modification experiments, we provided the mean position and orientation information of only the swimmers that have not yet reached the target in $s_{swimmers}$. The size of the state vector remains the same as in the previous case with performance curves for PPO and RPO as shown in figures \ref{fig:1c_ppo_ch6} and \ref{fig:1c_rpo_ch6} respectively. Here, we observe that convergence is faster and variance of return is less compared to the second set, but both PPO and RPO are stuck at local maxima. This is probably because steering is hard with just mean information. RPO shows relatively faster learning compared to PPO here as well.

Overall, we observe that performance reduces as the number of swimmers increases and state information reduces. This is likely because the agent needs more manoeuvres to get all the swimmers into the target which is of limited size. 

\begin{figure}
    \centering
    \begin{subfigure}{0.4\textwidth}
    \frame{\includegraphics[width=\textwidth]{ 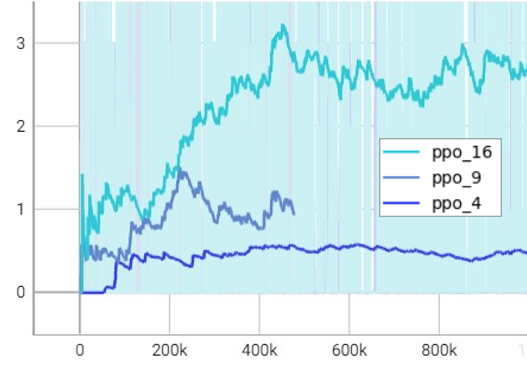}}
    \caption{}
    \label{fig:2_ppo_ch6}
    \end{subfigure}\hspace{0.5cm}
    \begin{subfigure}{0.4\textwidth}
    \centering
    \frame{\includegraphics[width=\textwidth]{ 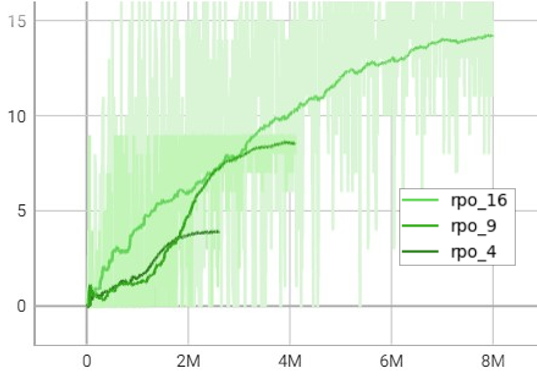}}
    \caption{}
    \label{fig:2_rpo_ch6}
    \end{subfigure}\hfill
    \caption{Performance of robustness experiments (random target location) with env-2 for (a) PPO and (b) RPO in terms of episodic returns with respect to time-steps for 4, 9 and 16 swimmers}
    \label{fig:random_target_ch6}
\end{figure}
    
\subsection{Robustness experiments}
Here, we performed experiments with target positions and size are random at every episode beginning. We constrained the target position to be within a 100um distance from the mean initial position of the swimmers along both x and y coordinate axes. The target radius is between 5 and 20um. Both the location and size of the target are chosen from a uniform distribution. The state information is taken to be like in the primary experiment, where positional and orientational information of all the swimmers are provided. Again, the experiment was performed for 4, 9 and 16 swimmers. Note that these experiments cover the cases of random mean initial position of swimmers also, where the positional state information can be centred to the mean initial position of swimmers without modifying the orientation information. Here, we observed the performance as shown in figure \ref{fig:random_target_ch6}. Here again, RPO performs extremely well compared to the PPO algorithm. PPO fails to converge for the simplest case of 4 swimmers, whereas, RPO can learn the navigation for 16 swimmers as well as observed in the subfigures \ref{fig:2_ppo_ch6} and \ref{fig:2_rpo_ch6} respectively. Here, we observe that the algorithm is burdened with two tasks of finding the path to the changing target as well as steering the swarm to get all the swimmers into the target. 

\begin{figure}
    \centering
    \frame{\includegraphics[width = 0.7\textwidth]{ 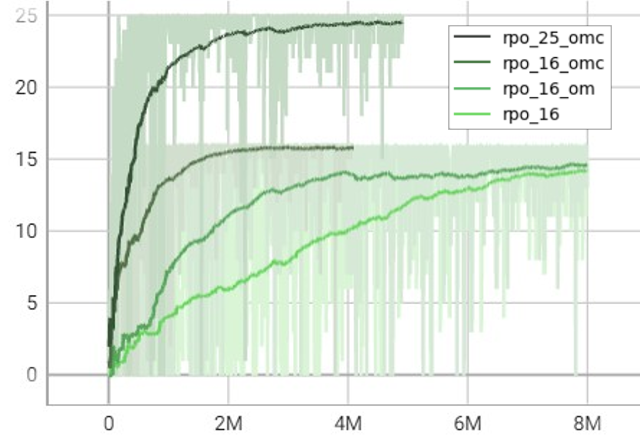}}
    \caption{Performance improvement in episodic return with orientation information, multiple environments and curriculum learning on the random target positon and size environment}
    \label{fig:env2star_ch6}
\end{figure}

We observed a slight improvement in performance when the target orientation with respect to the mean position of swimmers was provided in the state information. This experiment was performed for RPO only because of its consistently better performance compared to that of PPO and for 16 swimmers only as the result can be extrapolated for the simpler case of 4 and 9 swimmers as observed from the trend in the state information modification experiments. We also performed the sync vector environment with 4 parallel environments to obtain more data for the RL agent to learn effectively. The performance for this case is shown in figure \ref{fig:env2star_ch6} on the `$rpo\_16\_om$` line of the graph. 

We perform the final set of experiments incorporating the curriculum learning \cite{Bengio09} technique for the choice of the target position. The distance between the mean initial position of the swimmer swarm and the target was varied as shown in equation \ref{eqn:ch6_5}.

\begin{equation}
\label{eqn:ch6_5}
d \leq d_f - (d_f - d_s)*\exp^{-e_n/t_d}
\end{equation}

\begin{figure}[t]
    \centering
    \includegraphics[width=0.85\textwidth]{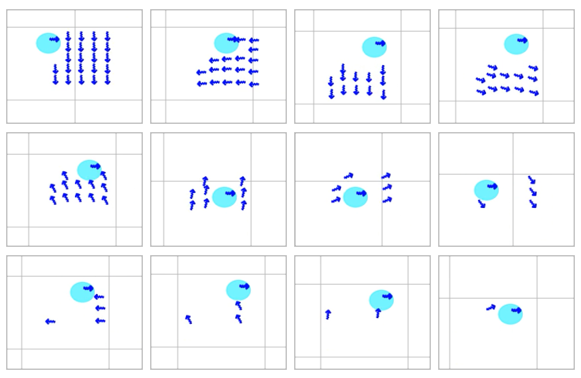}
    \caption{Navigation of 25 swimmers towards a given target}
    \label{fig:inference_ch6}
\end{figure}

where, $d_f$ is the final distance value, $d_s$ is the starting distance value, $e_n$ is the current episode number and $t_d$ is the threshold decay. We used 2 parallel environments to obtain more data in this case. This experiment was run for 16 swimmers with $t_d = 1000$ and 25 swimmers with $t_d = 2000$ and their performance is depicted in figure \ref{fig:env2star_ch6} on the `$rpo\_16\_omc$` and `$rpo\_25\_omc$` lines of the graph respectively. We observe that there is a tremendous improvement in the learning curve by using curriculum learning, which eases the first task of finding the target position and the agent gradually learns to manoeuvre the swarm around a target of limited size, especially observed in the 25 swimmers case.  

\subsection{RL model inference}
The RL agent that has been trained can be saved with the actor and critic model weights along with the running mean and variance of observation. A new environment once created will feed in the state information to the same model as during training but with updated weights. The trained agent now gives the orientation at each time step that the magnetic coil uses to align the micro-swimmers.

The set of images in figure \ref{fig:inference_ch6} show the steering performed by the RL agent for a swarm of 25 swimmers to reach a relatively smaller target. The arrow inside the target represents the swimmers absorbed by the target. 

\section{Conclusion}
\label{sec:conclusion_ch6}
Reinforcement learning proves effective in learning the navigation for a swarm of micro-swimmers with a single, global controller and in a complex experiment that is hard to model analytically. 

The RL agent finds it harder when number of swimmers increases or target size is smaller or target position is farther. It tends to converge at local minimum when the state information is reduced. It is also hard when there is randomness in the environment, where the target location is not constant. To improve the performance of navigation, more useful state information like the orientation of target with respect to the mean swimmer position can be provided and multiple environments can be run in parallel for more data as shown in the env-2-om experiment. Further, curriculum learning improves upon env-2-om where the agent is fed with easier (closer) targets at the beginning gradually moving towards tougher (farther) targets. Also, RPO has consistently shown better performance than PPO except for the scenario where the number of swimmers is 4, in which case PPO is achieving stable performance sooner than RPO. 

Future work could focus on reward modification to include distance-based metrics, which would produce non-sparse rewards.  Furthermore, the environment can be made more complex with obstacles and drifts, where the agent would be needed to be more robust to navigate towards the target.

\bibliographystyle{IEEEtran}

\bibliography{sn-article}

\begin{thebibliography}{10}
\providecommand{\url}[1]{#1}
\csname url@samestyle\endcsname
\providecommand{\newblock}{\relax}
\providecommand{\bibinfo}[2]{#2}
\providecommand{\BIBentrySTDinterwordspacing}{\spaceskip=0pt\relax}
\providecommand{\BIBentryALTinterwordstretchfactor}{4}
\providecommand{\BIBentryALTinterwordspacing}{\spaceskip=\fontdimen2\font plus
\BIBentryALTinterwordstretchfactor\fontdimen3\font minus
  \fontdimen4\font\relax}
\providecommand{\BIBforeignlanguage}[2]{{%
\expandafter\ifx\csname l@#1\endcsname\relax
\typeout{** WARNING: IEEEtran.bst: No hyphenation pattern has been}%
\typeout{** loaded for the language `#1'. Using the pattern for}%
\typeout{** the default language instead.}%
\else
\language=\csname l@#1\endcsname
\fi
#2}}
\providecommand{\BIBdecl}{\relax}
\BIBdecl

\bibitem{Li17}
\BIBentryALTinterwordspacing
J.~Li, B.~Esteban-Fern{\'a}ndez~de {\'A}vila, W.~Gao, L.~Zhang, and J.~Wang,
  ``Micro/nanorobots for biomedicine: Delivery, surgery, sensing, and
  detoxification,'' \emph{Sci Robot}, vol.~2, no.~4, Mar 2017, pMC6759331.
  [Online]. Available: \url{https://www.ncbi.nlm.nih.gov/pubmed/31552379}
\BIBentrySTDinterwordspacing

\bibitem{Goodrich17}
\BIBentryALTinterwordspacing
C.~P. Goodrich and M.~P. Brenner, ``Using active colloids as machines to weave
  and braid on the micrometer scale,'' \emph{Proceedings of the National
  Academy of Sciences}, vol. 114, no.~2, pp. 257--262, 2017. [Online].
  Available: \url{https://www.pnas.org/doi/abs/10.1073/pnas.1608838114}
\BIBentrySTDinterwordspacing

\bibitem{Yang20}
\BIBentryALTinterwordspacing
Y.~Yang and M.~A. Bevan, ``Cargo capture and transport by colloidal swarms,''
  \emph{Science Advances}, vol.~6, no.~4, p. eaay7679, 2020. [Online].
  Available: \url{https://www.science.org/doi/abs/10.1126/sciadv.aay7679}
\BIBentrySTDinterwordspacing

\bibitem{Soler13}
\BIBentryALTinterwordspacing
L.~Soler, V.~Magdanz, V.~M. Fomin, S.~Sanchez, and O.~G. Schmidt,
  ``Self-propelled micromotors for cleaning polluted water,'' \emph{ACS Nano},
  vol.~7, no.~11, pp. 9611--9620, Nov 2013. [Online]. Available:
  \url{https://doi.org/10.1021/nn405075d}
\BIBentrySTDinterwordspacing

\bibitem{Chen21}
\BIBentryALTinterwordspacing
L.~Chen, H.~Yuan, S.~Chen, C.~Zheng, X.~Wu, Z.~Li, C.~Liang, P.~Dai, Q.~Wang,
  X.~Ma, and X.~Yan, ``Cost-effective, high-yield production of biotemplated
  catalytic tubular micromotors as self-propelled microcleaners for water
  treatment,'' \emph{ACS Applied Materials {\&} Interfaces}, vol.~13, no.~26,
  pp. 31\,226--31\,235, Jul 2021. [Online]. Available:
  \url{https://doi.org/10.1021/acsami.1c03595}
\BIBentrySTDinterwordspacing

\bibitem{Li16RL}
\BIBentryALTinterwordspacing
J.~Li, I.~Rozen, and J.~Wang, ``Rocket science at the nanoscale,'' \emph{ACS
  Nano}, vol.~10, no.~6, pp. 5619--5634, Jun 2016. [Online]. Available:
  \url{https://doi.org/10.1021/acsnano.6b02518}
\BIBentrySTDinterwordspacing

\bibitem{Yin21}
\BIBentryALTinterwordspacing
C.~Yin, F.~Wei, S.~Fu, Z.~Zhai, Z.~Ge, L.~Yao, M.~Jiang, and M.~Liu, ``Visible
  light-driven jellyfish-like miniature swimming soft robot,'' \emph{ACS
  Applied Materials {\&} Interfaces}, vol.~13, no.~39, pp. 47\,147--47\,154,
  Oct 2021. [Online]. Available: \url{https://doi.org/10.1021/acsami.1c13975}
\BIBentrySTDinterwordspacing

\bibitem{Ider21}
A.~Daddi-Moussa-Ider, H.~Löwen, and B.~Liebchen, ``Hydrodynamics can determine
  the optimal route for microswimmer navigation,'' \emph{Communications
  Physics}, vol.~4, 02 2021.

\bibitem{Vinyals19}
\BIBentryALTinterwordspacing
O.~Vinyals, I.~Babuschkin, W.~M. Czarnecki, M.~Mathieu, A.~Dudzik, J.~Chung,
  D.~H. Choi, R.~Powell, T.~Ewalds, P.~Georgiev, J.~Oh, D.~Horgan, M.~Kroiss,
  I.~Danihelka, A.~Huang, L.~Sifre, T.~Cai, J.~P. Agapiou, M.~Jaderberg, A.~S.
  Vezhnevets, R.~Leblond, T.~Pohlen, V.~Dalibard, D.~Budden, Y.~Sulsky,
  J.~Molloy, T.~L. Paine, C.~Gulcehre, Z.~Wang, T.~Pfaff, Y.~Wu, R.~Ring,
  D.~Yogatama, D.~W{\"u}nsch, K.~McKinney, O.~Smith, T.~Schaul, T.~Lillicrap,
  K.~Kavukcuoglu, D.~Hassabis, C.~Apps, and D.~Silver, ``Grandmaster level in
  starcraft ii using multi-agent reinforcement learning,'' \emph{Nature}, vol.
  575, no. 7782, pp. 350--354, Nov 2019. [Online]. Available:
  \url{https://doi.org/10.1038/s41586-019-1724-z}
\BIBentrySTDinterwordspacing

\bibitem{Mnih15}
\BIBentryALTinterwordspacing
V.~Mnih, K.~Kavukcuoglu, D.~Silver, A.~A. Rusu, J.~Veness, M.~G. Bellemare,
  A.~Graves, M.~Riedmiller, A.~K. Fidjeland, G.~Ostrovski, S.~Petersen,
  C.~Beattie, A.~Sadik, I.~Antonoglou, H.~King, D.~Kumaran, D.~Wierstra,
  S.~Legg, and D.~Hassabis, ``Human-level control through deep reinforcement
  learning,'' \emph{Nature}, vol. 518, no. 7540, pp. 529--533, Feb 2015.
  [Online]. Available: \url{https://doi.org/10.1038/nature14236}
\BIBentrySTDinterwordspacing

\bibitem{Kober13}
J.~Kober, J.~Bagnell, and J.~Peters, ``Reinforcement learning in robotics: A
  survey,'' \emph{The International Journal of Robotics Research}, vol.~32, pp.
  1238--1274, 09 2013.

\bibitem{breyer19}
M.~Breyer, F.~Furrer, T.~Novkovic, R.~Siegwart, and J.~Nieto, ``Comparing task
  simplifications to learn closed-loop object picking using deep reinforcement
  learning,'' 2019.

\bibitem{Kul19}
J.~Kulhánek, E.~Derner, T.~de~Bruin, and R.~Babuška, ``Vision-based
  navigation using deep reinforcement learning,'' in \emph{2019 European
  Conference on Mobile Robots (ECMR)}, 2019, pp. 1--8.

\bibitem{kalashnikov18qtopt}
D.~Kalashnikov, A.~Irpan, P.~Pastor, J.~Ibarz, A.~Herzog, E.~Jang, D.~Quillen,
  E.~Holly, M.~Kalakrishnan, V.~Vanhoucke, and S.~Levine, ``Qt-opt: Scalable
  deep reinforcement learning for vision-based robotic manipulation,'' 2018.

\bibitem{Waschneck18}
B.~Waschneck, A.~Reichstaller, L.~Belzner, T.~Altenmüller, T.~Bauernhansl,
  A.~Knapp, and A.~Kyek, ``Deep reinforcement learning for semiconductor
  production scheduling,'' in \emph{2018 29th Annual SEMI Advanced
  Semiconductor Manufacturing Conference (ASMC)}, 2018, pp. 301--306.

\bibitem{liu21finrl}
X.-Y. Liu, H.~Yang, J.~Gao, and C.~D. Wang, ``{FinRL}: Deep reinforcement
  learning framework to automate trading in quantitative finance,'' \emph{ACM
  International Conference on AI in Finance (ICAIF)}, 2021.

\bibitem{Yu21}
\BIBentryALTinterwordspacing
C.~Yu, J.~Liu, S.~Nemati, and G.~Yin, ``Reinforcement learning in healthcare: A
  survey,'' \emph{ACM Comput. Surv.}, vol.~55, no.~1, nov 2021. [Online].
  Available: \url{https://doi.org/10.1145/3477600}
\BIBentrySTDinterwordspacing

\bibitem{Schulman17}
J.~Schulman, F.~Wolski, P.~Dhariwal, A.~Radford, and O.~Klimov, ``Proximal
  policy optimization algorithms,'' 2017.

\bibitem{Rahman23}
\BIBentryALTinterwordspacing
M.~M. Rahman and Y.~Xue, ``Robust policy optimization in deep reinforcement
  learning,'' 2022. [Online]. Available:
  \url{https://openreview.net/forum?id=HnLFY8F9uS}
\BIBentrySTDinterwordspacing

\bibitem{Yang19}
Y.~Yang, M.~Bevan, and B.~Li, ``Efficient navigation of colloidal robots in an
  unknown environment via deep reinforcement learning,'' \emph{Advanced
  Intelligent Systems}, vol.~2, 09 2019.

\bibitem{Falk21}
\BIBentryALTinterwordspacing
M.~J. Falk, V.~Alizadehyazdi, H.~Jaeger, and A.~Murugan, ``Learning to control
  active matter,'' \emph{Phys. Rev. Res.}, vol.~3, p. 033291, Sep 2021.
  [Online]. Available:
  \url{https://link.aps.org/doi/10.1103/PhysRevResearch.3.033291}
\BIBentrySTDinterwordspacing

\bibitem{Muinos21}
S.~Muiños-Landin, A.~Fischer, V.~Holubec, and F.~Cichos, ``Reinforcement
  learning with artificial microswimmers,'' \emph{Science Robotics}, vol.~6,
  no.~52, p. eabd9285, 2021.

\bibitem{Yang20RL}
\BIBentryALTinterwordspacing
Y.~Yang, M.~A. Bevan, and B.~Li, ``Micro/nano motor navigation and localization
  via deep reinforcement learning,'' \emph{Advanced Theory and Simulations},
  vol.~3, no.~6, p. 2000034, 2020. [Online]. Available:
  \url{https://onlinelibrary.wiley.com/doi/abs/10.1002/adts.202000034}
\BIBentrySTDinterwordspacing

\bibitem{Ghosh09}
\BIBentryALTinterwordspacing
A.~Ghosh and P.~Fischer, ``Controlled propulsion of artificial magnetic
  nanostructured propellers,'' \emph{Nano Letters}, vol.~9, no.~6, pp.
  2243--2245, Jun 2009. [Online]. Available:
  \url{https://doi.org/10.1021/nl900186w}
\BIBentrySTDinterwordspacing

\bibitem{Pal22}
\BIBentryALTinterwordspacing
M.~Pal, I.~Fouxon, A.~M. Leshansky, and A.~Ghosh, ``Fluid flow induced by
  helical microswimmers in bulk and near walls,'' \emph{Phys. Rev. Res.},
  vol.~4, p. 033069, Jul 2022. [Online]. Available:
  \url{https://link.aps.org/doi/10.1103/PhysRevResearch.4.033069}
\BIBentrySTDinterwordspacing

\bibitem{Bechinger16}
\BIBentryALTinterwordspacing
C.~Bechinger, R.~Di~Leonardo, H.~L\""owen, C.~Reichhardt, G.~Volpe, and
  G.~Volpe, ``Active particles in complex and crowded environments,''
  \emph{Rev. Mod. Phys.}, vol.~88, p. 045006, Nov 2016. [Online]. Available:
  \url{https://link.aps.org/doi/10.1103/RevModPhys.88.045006}
\BIBentrySTDinterwordspacing

\bibitem{Purcell77}
\BIBentryALTinterwordspacing
E.~M. Purcell, ``{Life at low Reynolds number},'' \emph{American Journal of
  Physics}, vol.~45, no.~1, pp. 3--11, 01 1977. [Online]. Available:
  \url{https://doi.org/10.1119/1.10903}
\BIBentrySTDinterwordspacing

\bibitem{Schulman15}
\BIBentryALTinterwordspacing
J.~Schulman, S.~Levine, P.~Abbeel, M.~Jordan, and P.~Moritz, ``Trust region
  policy optimization,'' in \emph{Proceedings of the 32nd International
  Conference on Machine Learning}, ser. Proceedings of Machine Learning
  Research, F.~Bach and D.~Blei, Eds., vol.~37.\hskip 1em plus 0.5em minus
  0.4em\relax Lille, France: PMLR, 07--09 Jul 2015, pp. 1889--1897. [Online].
  Available: \url{https://proceedings.mlr.press/v37/schulman15.html}
\BIBentrySTDinterwordspacing

\bibitem{Huang22}
\BIBentryALTinterwordspacing
S.~Huang, R.~F.~J. Dossa, C.~Ye, J.~Braga, D.~Chakraborty, K.~Mehta, and J.~G.
  AraÃºjo, ``Cleanrl: High-quality single-file implementations of deep
  reinforcement learning algorithms,'' \emph{Journal of Machine Learning
  Research}, vol.~23, no. 274, pp. 1--18, 2022. [Online]. Available:
  \url{http://jmlr.org/papers/v23/21-1342.html}
\BIBentrySTDinterwordspacing

\bibitem{Bengio09}
\BIBentryALTinterwordspacing
Y.~Bengio, J.~Louradour, R.~Collobert, and J.~Weston, ``Curriculum learning,''
  in \emph{Proceedings of the 26th Annual International Conference on Machine
  Learning}, ser. ICML '09.\hskip 1em plus 0.5em minus 0.4em\relax New York,
  NY, USA: Association for Computing Machinery, 2009, p. 41–48. [Online].
  Available: \url{https://doi.org/10.1145/1553374.1553380}
\BIBentrySTDinterwordspacing

\end{thebibliography}

\end{document}